%% file: main.tex
\documentclass[10pt,twocolumn,letterpaper]{article}

\usepackage{iccv}              % To produce the CAMERA-READY version
\usepackage{graphicx}
\usepackage{amsmath}
\usepackage{amssymb}
\usepackage{pifont}
\usepackage{booktabs}
\usepackage{float}
\usepackage{algorithm}
\usepackage{algpseudocode}
\usepackage{multicol} % For multi-column formatting
\usepackage{adjustbox} % To adjust table width
\usepackage{multirow}

\definecolor{iccvblue}{rgb}{0.21,0.49,0.74}
\usepackage[pagebackref,breaklinks,colorlinks,allcolors=iccvblue]{hyperref}

\def\paperID{8613} % *** Enter the Paper ID here
\def\confName{ICCV}
\def\confYear{2025}
 
\title{SPG: Improving Motion Diffusion by Smooth Perturbation Guidance
}

\author{Boseong Jeon\\
Samsung Research\\
Republic of Korea \\
{\tt\small bf.jeon@samsung.com}
}
\begin{document}  

\maketitle

\begin{figure*}
\centering
\includegraphics[width=0.86\linewidth]{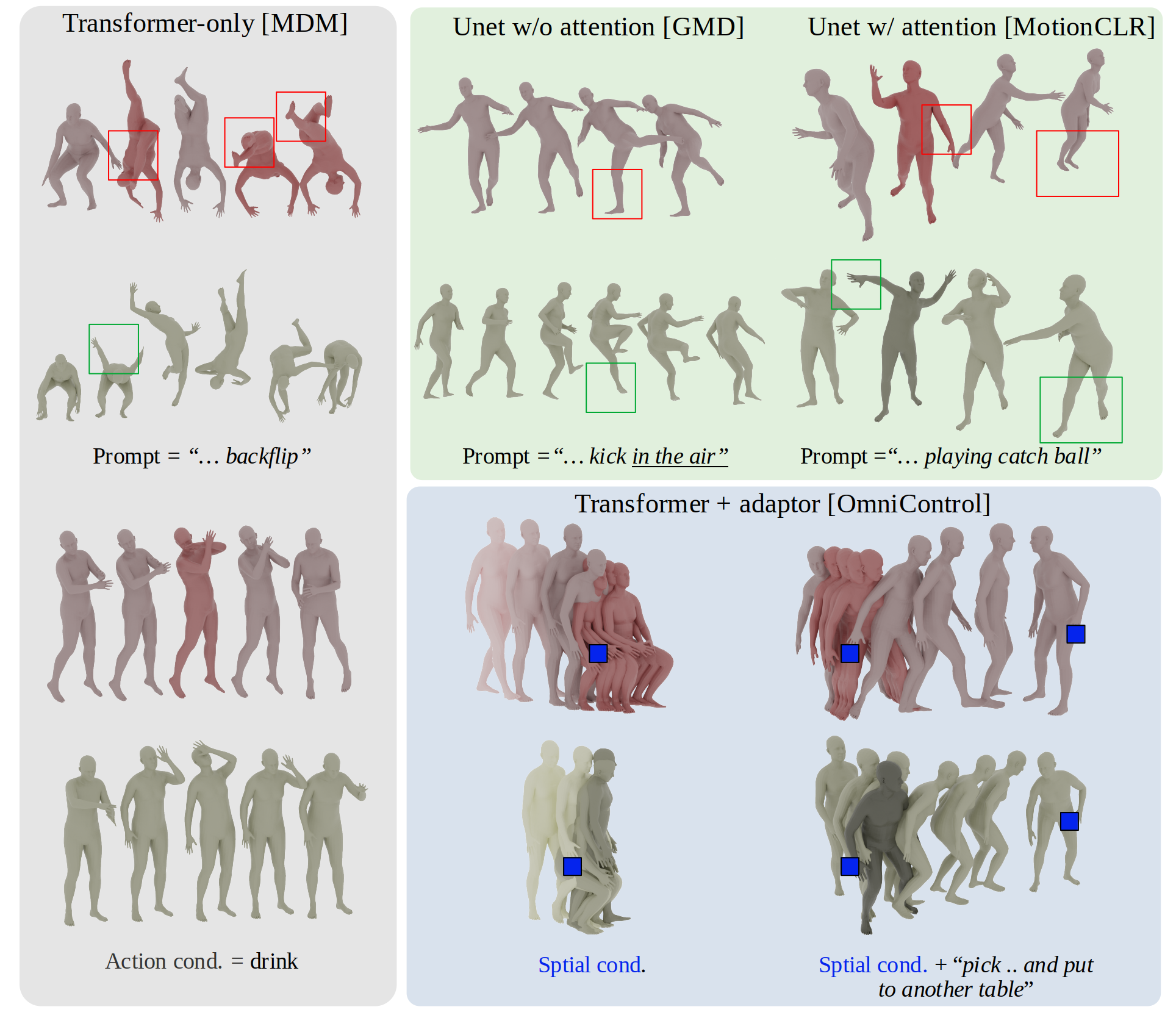}
   \caption{\textcolor{red}{Red}: baseline (CFG or no guidance), \textcolor{green}{Green}: only SPG applied. With fewer than 10 lines of code changes during test-time inference, SPG improves the realism and reduces foot-skating in motion diffusion models. SPG is a model-agnostic weak-model guidance approach applicable to various networks and tasks.}

\label{fig:long}
\label{fig:onecol}
\end{figure*}

%%%%%%%%% ABSTRACT
\begin{abstract}

This paper presents a test-time guidance method to improve the output quality of the human motion diffusion models without requiring additional training.
To have negative guidance, Smooth Perturbation Guidance (SPG) builds a weak model by temporally smoothing the motion in the denoising steps. Compared to model-agnostic methods originating from the image generation field, SPG effectively mitigates out-of-distribution issues when perturbing motion diffusion models. In SPG guidance, the nature of motion structure remains intact.
This work conducts a comprehensive analysis across distinct model architectures and tasks. Despite its extremely simple implementation and no need for additional training requirements, SPG consistently enhances motion fidelity. Project page can be found at \url{https://spg-public.vercel.app/}.

\end{abstract}

%%%%%%%%% BODY TEXT

\section{Introduction}

This paper explores how to enhance the test-time denoising process in various motion diffusion models \cite{mdm, omnicontrol, motionclr, gmd, dabral2023mofusion} with different network structures. Previously, motion diffusion models relied on either classifier guidance (CG) \cite{cg} or classifier-free guidance (CFG) \cite{cfg} to improve performance.
While CG in motion diffusion models \cite{omnicontrol, gmd, zhang2024scenic, wei2024diffkfc} can be expressed mathematically, it is prone to local minima and requires manual tuning of the number of iterations. On the other hand, CFG has been widely adopted in motion diffusion models \cite{mdm, remofussion, gmd, mld, shafir2023priormdm}, but it demands additional training and cannot be applied in unconditional settings \cite{raab2023modi}. Moreover, studies have reported that CFG reduces generation diversity \cite{autoguidance, pag}.

In the field of image and video diffusion models, several studies \cite{sg, pag, sag, icg, hyung2024spatiotemporal, hong2024seg} have explored using weak model outputs to replace the negative term in CFG, demonstrating effectiveness within the art community \cite{sdart}. Instead of requiring additional training, these approaches utilize \textit{aligned weak} outputs as negative guidance by perturbing either the inputs or intermediate activations of the network, which requires only minimal code modifications during inference.

Despite the success of guidance techniques in other domains, a systematic analysis of weak model guidance in motion generation remains limited. 
% To the best of our knowledge, only the authors of \cite{icg} included a motion diffusion (MDM \cite{mdm}) in their results. However, the main focus of the paper was image generation and did not include a detailed analysis nor applicability to various motion diffusions. 
Perturbing motion diffusion outputs without introducing out-of-distribution issues \cite{sg, pag} is particularly challenging due to the scarcity of data  and the complexity of the state manifold \cite{huang2024constraineddiffusiontrustsampling}. 
This paper aims to bridge this gap by investigating whether the weak model guidance techniques can be effectively applied to motion synthesis while considering the unique characteristics of motion generation.
To this end, we propose Smooth Perturbation Guidance (SPG), a method that leverages a simple yet effective insight: smoothing motion trajectories along the temporal axis reduces motion details while preserving overall structure, making them useful as aligned weaker outputs. 

We validate SPG across various architectures, including networks without self-attention \cite{gmd} and those incorporating ControlNet-like guidance mechanisms \cite{omnicontrol}. Additionally, we evaluate SPG on multiple motion generation tasks, including text-to-motion (T2M), action-to-motion (A2M), unconstrained generation, and trajectory following.
Compared to other weak model guidance methods originally developed for image generation, SPG achieves state-of-the-art performance. Compared to CFG, SPG offers comparable performance while improving diversity. Moreover, combining SPG with CFG yields higher  fidelity in most cases, demonstrating their complementary nature.
To the best of our knowledge, this work represents the first systematic effort to develop a weak model guidance approach specifically tailored for motion generation.

\section{Related Work}
\subsection{Motion Diffusion Model}
\label{sec: Motion Diffusion Model}
Human motion synthesis synthesizes  joint trajectories  based on conditions like text \cite{mdm, humanml3d, shafir2023priormdm, remofussion, zhang2022motiondiffuse}, action \cite{petrovich2021action, guo2020action2motion}, music \cite{aristidou2022rhythm, alexanderson2023listen}, reference trajectories \cite{omnicontrol, gmd}, and incomplete motions \cite{condmdi, zhang2024rohm, agrol, wei2024diffkfc}. Similar to video generation \cite{hyung2024spatiotemporal, nvidia2025cosmosworldfoundationmodel}, motion can be generated either in an auto-regressive manner \cite{qin2022motion, jiang2023motiongpt, li2024aamdm}, or as fixed-length sequences using diffusion models \cite{mdm, omnicontrol, motionclr, shafir2023priormdm}. Thanks to its superior global coherence and stable diversity, the latter approach has emerged as a promising method for motion generation.
In motion diffusion models, architectures vary significantly rather than having a common backbone (e.g., 2D \cite{sd, podell2023sdxl} and 3D UNet \cite{sdvideo} with attention blocks or DiT \cite{dit, esser2024scalingrectifiedflowtransformers}) , which are more standardized in image and video diffusion \cite{sd, sdvideo, podell2023sdxl, xing2023dynamicrafteranimatingopendomainimages}. In the image generation domain, there is a group of works \cite{wang2024instantstyle, frenkel2024implicit, jeong2024visual} which bootstrap and harness the property of the specific backbone structures. 

In the motion domain, in contrast, there are few works in a similar direction. The diffusion networks in \cite{mdm, zhang2024rohm} employ a transformer-encoder blocks. On the other hand, the networks in \cite{gmd, condmdi, zhao2023modiff} adopt a UNet architecture without self-attention, while \cite{motionclr, raab2023single} utilizes a UNet with cross-attention or self-attention. Meanwhile, motion prediction networks such as \cite{agrol} use a simple MLP-only architecture.

To enhance sampling quality and achieve the desired behaviors, CG and CFG are utilized for motion diffusion models. CFG is specifically employed to improve text alignment \cite{mld, mdm, remofussion} or blending motions \cite{shafir2023priormdm}.
However, implementing CFG necessitates additional training on empty prompts or specific conditions, and finding the right balance between fidelity and diversity is non-trivial \cite{remofussion}. Moreover, CFG may be ineffective or unavailable \cite{mdm, raab2023modi}. For instance, MDM \cite{mdm} does not provide CFG-trained models for unconditional or action-based tasks. 
To accommodate more sophisticated tasks, such as trajectory following and obstacle avoidance, CG is utilized \cite{omnicontrol, gmd, huang2024constraineddiffusiontrustsampling} by adjusting the denoising direction based on the gradients of performance measures.
However, these approaches are prone to local minima and require careful tuning of the number of iterations for gradient computation, making them challenging to implement universally.

\subsection{Guidance with weak model}
\label{sec:Guidance with weak model}
In the fields of image and video generation, researchers have proposed weak models \cite{icg, sg, autoguidance, pag,  hyung2024spatiotemporal, hong2024seg} that compute a lower-quality model output for extending the unconditional term of CFG into a more general form of negative guidance, which does not rely on training with the empty prompt.
These approaches generally fall into two categories: (1) those that perturb intermediate network outputs by leveraging an understanding of the model’s internal structure \cite{pag, autoguidance, hyung2024spatiotemporal}, and (2) those that perturb inputs to the denoising network \cite{sag, sg, icg} in a model-agnostic manner. In the first category, methods like PAG \cite{pag} and SEG \cite{hong2024seg} replace cross-attention of specific blocks with an identity matrix or blur the attention weights. 
STG \cite{hyung2024spatiotemporal} skips certain layers, and AutoGuidance\cite{autoguidance} introduces dropout techniques or uses less-trained networks. 

The second category does not depend on the model’s architecture and instead perturbs specific inputs to the denoising network. SAG \cite{sag} applies Gaussian blurring to the state input channel (note that SAG can be used without self-attention if skipping the region selection for blurring), SG \cite{sg} alters timesteps to obtain a more dispersed distribution in higher noise levels, and ICG \cite{icg} modifies the conditional embeddings.
This work aims to develop the latter model-agnostic approach to accommodate motion diffusion models without depending on a specific network structure.   

Compared to CG and CFG, these weak-model guidance methods offer several advantages. They enhance fidelity while maintaining better diversity. Additionally, they can be applied in unconditional settings, expanding their usability across different tasks. Furthermore, the weak term can be combined with CG \cite{sag}, CFG \cite{pag}, or even other weak models \cite{sg}, providing greater flexibility in guiding the generation process.

\subsection{Designing aligned weak model}
\label{sec: Designing aligned weak model}
To build an effective weak model, it's essential to ensure alignment between the model's output and its perturbation by degrading only the details while preserving the core content.
For instance, the authors of PAG \cite{pag} addressed the out-of-distribution issue in perturbations by manipulating cross-attention instead of modifying the inputs. This approach allows fine structures to collapse while maintaining the overall appearance. Similarly, SAG \cite{sag} employs blur-guidance to degrade only the fine details, ensuring that the overall structure of the original information remains intact.
The authors of STG \cite{hyung2024spatiotemporal} also highlighted the importance of \textit{slightly} weak model for their layer-skipping design.

Compared to image generation, ensuring  alignment in weak models for motion diffusion presents unique challenges. One major factor is the significant difference in the scale and diversity of training datasets between the image and motion domains. For example, image generation models like Stable Diffusion \cite{sd} are trained on billion-scale datasets such as LAION \cite{schuhmann2022laion}, whereas commonly used motion datasets \cite{humanml3d, kit} collectively contain fewer than 100K samples. This limited data makes it difficult for motion diffusion networks to ensure reliable perturbations, particularly in underrepresented regions of the motion data space.

Another challenge arises from the representation differences between motion and image diffusion models. While image and video diffusion models typically operate in latent space \cite{autoencoderkl, vldm}, many motion diffusion models \cite{mdm, omnicontrol, gmd, motionclr, wei2024diffkfc, yi2024generatinghumaninteractionmotions, qin2022motion, liu2023plan, raab2023single} rely on position or joint representations \cite{humanml3d}. These representations vary significantly in scale across different dimensions and exist on a sparser manifold compared to their image-based counterparts \cite{gmd}. As a result, naively applying perturbations can easily lead to unexpected behaviors, making it more challenging to design a robust weak model.

\section{Preliminaries}
\subsection{Motion diffusion models}

Given a motion dataset $\mathbf{x}_0 \sim q(\mathbf{x}_0)$, motion diffusion models iteratively add Gaussian noise to the samples over $T$ steps. This process ensures that the marginal distribution at step $T$ approximates a standard Gaussian, i.e., $q(\mathbf{x}_T) \approx \mathcal{N}(\mathbf{0}, \mathbf{I})$. The forward diffusion process is defined as:

\[
{
q(\mathbf{x}_{t} | \mathbf{x}_{t-1}) = \mathcal{N}(\sqrt{\alpha_{t}} \mathbf{x}_{t-1},(1-\alpha_{t})\mathbf{I}),
}
\]
where $t$ represents the diffusion step, and $\alpha_t$ is a predefined variance schedule that controls the amount of noise added at each step. 
$\mathbf{x}_t$ corresponds to the trajectory of the state $\mathbf{x}_t \in \mathbb{R}^{J\times N}$ with the spatial dimension $J$ and the temporal length $N$. It is based on the state representation of \cite{humanml3d}, described in either  local \cite{mdm} or global \cite{gmd} coordinates. 

To generate samples conditioned on a given condition $c$, diffusion models learn the reverse process, which involves gradually removing noise from $\mathbf{x}_t$, starting from pure Gaussian noise $\mathbf{x}_T$. Given model parameters $\theta$, the reverse process is modeled as:
\begin{equation}
    p_\theta(\mathbf{x}_{t-1} | \mathbf{x}_t, c) := \mathcal{N}(\mathbf{x}_{t-1}; \mu_\theta(\mathbf{x}_t,c), \Sigma_t),
    \label{eqn: denosing}
\end{equation}
where $\Sigma_t = \frac{1 - {\bar{\alpha}_{t-1}} \beta_t}{1 - \bar{\alpha_t}} \mathbf{I}$ and $\beta_t=1-\alpha_t$. For the simplicity of loss design \cite{mdm, omnicontrol, wei2024diffkfc}, the majority of motion diffusion models $g_{\theta}(\mathbf{x}_t,  c)$ are formulated as the clean prediction $\hat{\mathbf{x}}_0$ at step $t$. Thus, we can compute $\mu_\theta(\mathbf{x}_t,c)$ as:

\[
{
 \mu_\theta(\mathbf{x}_t,c) = \frac{\sqrt{\bar{\alpha}_{t-1}} \beta_t}{1 - \bar{\alpha_t}} g_{\theta}(\mathbf{x}_t, c) + \frac{\sqrt{\alpha_t} (1 - \bar{\alpha}_{t-1})}{1 - \bar{\alpha}_t} \mathbf{x}_t,
}
\]
with $\bar{\alpha_t} := \prod_{s=1}^{t} \alpha_s$. For brevity, $t$ is omitted from the inputs of $\mu_\theta(\cdot)$ and $g_{\theta}(\cdot)$.

\subsection{Guidances in diffusion models}

In the general form \cite{cfg, cg}, guidance in the denoising process (\ref{eqn: denosing}) is achieved by adding the gradient of a given cost function, \( J(\mathbf{x}_t, c) \), to the score function \( g_{\theta}(\mathbf{x}_t, c) \):

\[
 g_{\theta}(\mathbf{x}_t,  c) + s \nabla_{\mathbf{x}_t} J(\mathbf{x}_t, c).
\]
In classifier-free guidance (CFG), the cost function is set as:
\[
J(\mathbf{x}_t, c) = -\log \frac{p_\theta(\mathbf{x}_t | c)}{p_{\theta}(\mathbf{x}_t)},
\]
which increases the ratio of the conditional probability to the unconditional probability. Diffusion models can utilize CFG-based sampling to enhance prompt alignment and output quality using a positive scale parameter \( s \):

\begin{equation}
\label{eqn: cfg}
\hat{\mathbf{x}}_{0}= g_{\theta}(\mathbf{x}_t, c)  + s \big(g_{\theta}(\mathbf{x}_t, c) - g_{\theta}(\mathbf{x}_t, \phi) \big),
\end{equation}
where \( g_{\theta}(\mathbf{x}_t, \phi) \) is obtained by training with an empty prompt \( \phi \) using a specified drop probability \cite{mdm, condmdi}.

Although not exactly the same, most weak model guidance methods \cite{pag, sg, sag, autoguidance, icg} extend the CFG framework by conceptually defining the cost based on the ratio of good samples to bad samples:
\[
J(\mathbf{x}_t, c) = -\log \frac{p_\theta(\mathbf{x}_t | c)}{p_\theta(\mathbf{x}_t | \Tilde{c})},
\]
where \( \Tilde{c} \) represents a class corresponding to poor-quality samples. Equation (\ref{eqn: cfg}) can be extended as follows, with a score function corresponding to the weak model output classified into $\Tilde{c}$:

\begin{equation}
\label{eqn: weak}
\hat{\mathbf{x}}_{0}= g_{\theta}(\mathbf{x}_t, c)  + s \big(g_{\theta}(\mathbf{x}_t, c) - \Tilde{g}_{\theta}(\mathbf{x}_t, c) \big).
\end{equation}
Here, we refer to \( \Tilde{g}_{\theta}(\mathbf{x}_t, c) \) as the \textit{weak term} which is  a degraded version of $c$ and classified into \( \Tilde{c} \). Without additional training, the weak term can be built by perturbing the original model output \( g_{\theta}(\mathbf{x}_t, c) \) similarly with methods introduced in Section \ref{sec:Guidance with weak model}.

\begin{figure}[!t]
\centering
\includegraphics[width=0.7\linewidth]{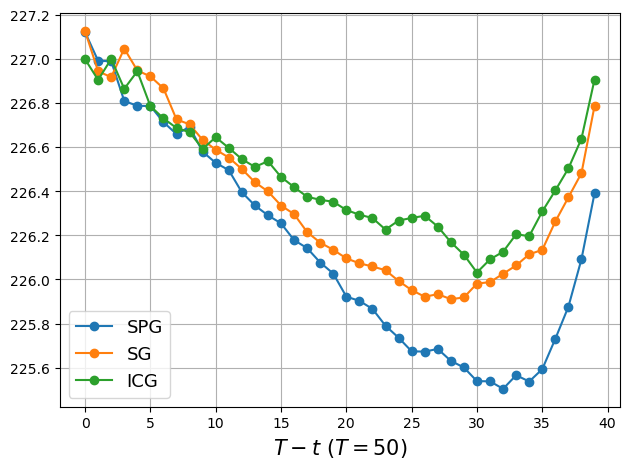}
   \caption{An estimate of how the sampling process deviates from the state manifold \cite{huang2024constraineddiffusiontrustsampling}. A lower value of $\|\epsilon(\mathbf{x}_t, c)\|$ indicates better confinement within the state manifold. Batch size was set to $10$.}

\label{fig:manifold}
\end{figure}

\begin{figure*}[!t]
\centering
\includegraphics[width=0.96\linewidth]{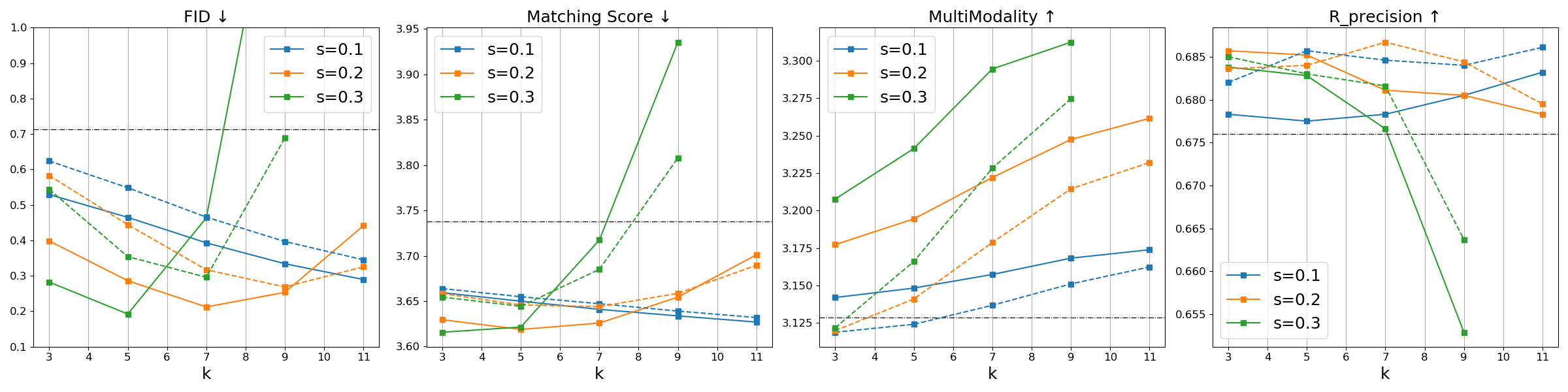}
\includegraphics[width=0.96\linewidth]{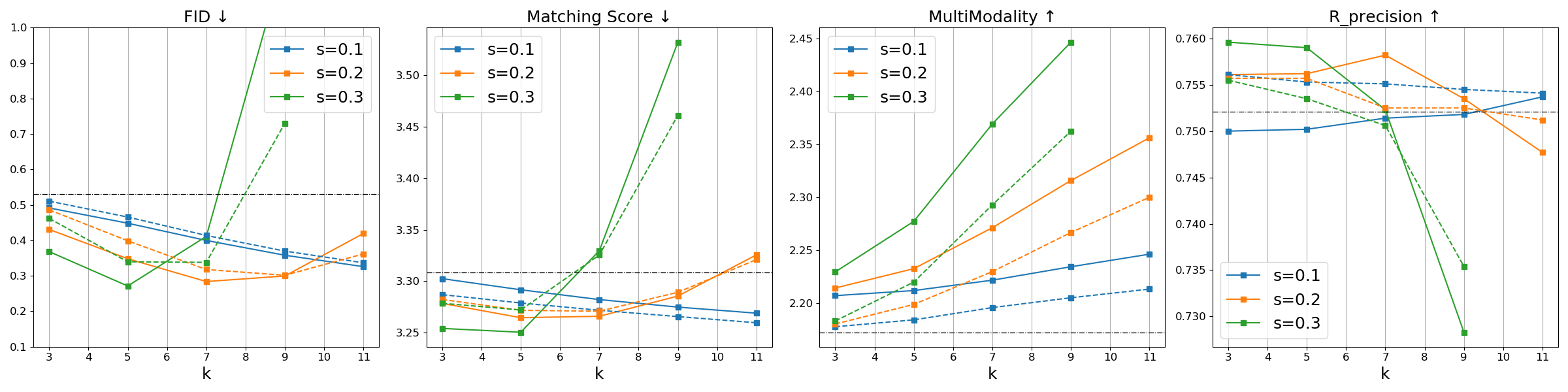}
\caption{Performance for various SPG scales $s$ in (\ref{eqn: spg}) and kernel sizes $k$ in (\ref{eqn: convolution}). Evaluations were conducted on the HumanML3D test set using the T2M model of MDM. (Top) without CFG, (Bottom) with CFG. Black dash denotes the baseline without any weak model guidnaces.
Colored dashed lines were obtained from the original SAG \cite{sag} implementation on deterministic noise in equation (\ref{eqn: diffusing from smooth}) while solid lines correspond to SPG. With a proper scale $k \leq 7$ and $s \geq 0.2$, SPG achieved better result than the baseline for the most metrics. Best viewed in color.} 
\label{fig: k scale}
\end{figure*}

\section{SPG: Smooth Perturbation Guidance}

\subsection{Approach}
This section introduces SPG as a method for constructing a model-agnostic, aligned weak model, taking into account the susceptibility of motion diffusion models to perturbations and its applicability to models without attention mechanisms \cite{gmd, condmdi}.
As mentioned in Section \ref{sec: Designing aligned weak model}, randomly perturbing the inputs of the motion denoising network often results in out-of-distribution data. To address this, we need a technique that ensures the perturbed inputs remain within the valid motion manifold.

We generate the weak version of $g_{\theta}(\mathbf{x}_t, c)$ by applying smoothing to $\mathbf{x}_t$ only \textbf{along the temporal axis}. This approach offers several advantages. First, as discussed in SAG \cite{sag}, applying Gaussian blurring as a perturbation can selectively degrade fine details while preserving the overall structure when applied with an appropriate blur scale.
Also, the temporally perturbed motion is more likely to be a valid human motion, as it preserves the natural temporal continuity of movement. In contrast, applying spatial smoothing can lead to unrealistic motion patterns; for example, simultaneously smoothing both hands and feet can result in unnatural or physically implausible movements. 
Thus, SPG selectively degrades fine details while preserving the structural integrity of human motion. 
The comparison of applying blur to both the spatial and temporal axes is provided in the supplementary material.

% \begin{algorithm}[t!]
%     \caption{Smooth Perturbation Guidance (SPG)}
%     \label{alg:spg}
%     \begin{algorithmic}[1]
%         \Require Motion model $g_{\theta}$, input state $\mathbf{x}_t$, condition $c$, window size $k$, scaling factor $s$
%         \Ensure Refined prediction $\hat{\mathbf{x}}_{0}$
        
%         \State \textbf{Compute Predicted Motion:} 
%         \State $\hat{\mathbf{x}}_{0} \gets g_{\theta}(\mathbf{x}_t, c)$ 
        
%         \State \textbf{Apply Temporal Smoothing:}
%         \State Define kernel $W_k \gets \frac{1}{k} (1,1, \dots, 1)$ 
%         \State $\Tilde{\mathbf{x}}_{0} \gets \hat{\mathbf{x}}_{0} * W_k$
        
%         \State \textbf{Diffuse to Noise Level $t$:}
%         \State Sample $\mathbf{\epsilon} \sim \mathcal{N}(0, I)$
%         \State $\Tilde{\mathbf{x}}_{t} \gets \sqrt{\bar{\alpha}_{t}} \Tilde{\mathbf{x}}_{0} + \sqrt{1 - \bar{\alpha}_{t}} \mathbf{\epsilon}$
        
%         \State \textbf{Apply Smooth Perturbation Guidance:}
%         \State $\hat{\mathbf{x}}_{0} \gets g_{\theta}(\mathbf{x}_t, c) + s \big(g_{\theta}(\mathbf{x}_t, c) - g_{\theta}(\Tilde{\mathbf{x}}_{t}, c)\big)$
        
%         \State \textbf{Return} $\hat{\mathbf{x}}_{0}$
%     \end{algorithmic}
% \end{algorithm}

\subsection{Method}
Given the motion model's prediction at timestep $t$, denoted as $\hat{\mathbf{x}}_{0} = g_{\theta}(\mathbf{x}_t, c)$, a smoothed trajectory $\Tilde{\mathbf{x}}_{0}$ can be obtained using a simple moving average with a window size of $k$, formulated as:

\begin{equation}
\label{eqn: convolution}
\Tilde{\mathbf{x}}_{0} = \hat{\mathbf{x}}_{0} * W_k,
\end{equation}

where the kernel $W_k$ is defined as $W_k = \frac{1}{k} (1, 1, \dots, 1)$.
By diffusing the perturbed input $\Tilde{\mathbf{x}}_{0}$ to noise level $t$, we obtain:
\begin{equation}
\label{eqn: diffusing from smooth}
\Tilde{\mathbf{x}}_{t} = \sqrt{\bar{\alpha_{t}}} \Tilde{\mathbf{x}}_{0} + \sqrt{1 - \bar{\alpha_{t}}} \mathbf{\epsilon},
\end{equation}
where $\epsilon \sim \mathcal{N}(0, I)$. Unlike SAG, which employs a deterministic $\epsilon$ computed from $(\mathbf{x}_t, \hat{\mathbf{x}}_{0})$, we found that using randomly sampled Gaussian noise yields superior performance. The corresponding ablation results are presented in Figure \ref{fig: k scale}.
Using the smoothed input $\Tilde{\mathbf{x}}_{t}$ at noise level $t$, SPG is formulated as follows:
\begin{equation}
\label{eqn: spg}
\hat{\mathbf{x}}_{0}  = g_{\theta}(\mathbf{x}_t, c) +  s \big(g_{\theta}(\mathbf{x}_t, c) - g_{\theta}(\Tilde{\mathbf{x}}_{t}, c)\big),
\end{equation}
where $s$ is a positive scaling factor. The effect of varying the guidance scale $s$ and kernel size $k$ is depicted in Figure \ref{fig: k scale}.

\begin{table}[!t]
    \centering
    \begin{tabular}{lcccc}
    \toprule
     & FID ↓ & Matching Score ↓ & R precision ↑ \\
    \midrule
    CFG & \textbf{3.127} & 4.699 & 0.552 \\
    ICG & 3.791 & {4.061} & {0.645} \\
    SPG & {3.235} & \textbf{3.833} & \textbf{0.677} \\
    \bottomrule
    \end{tabular}  
    \\
    \caption{Performance metric of the weak term in the early denoising step.}
    \label{tab: Alignment comparison}
\end{table}

\subsection{Analysis}
By introducing the temporal smoothing $\Tilde{\mathbf{x}}_{t}$, the perturbed model output remains within the motion dataset manifold more effectively than arbitrary perturbations, such as adding random noise to the time embedding, as done in ICG \cite{icg}. Based on MDM model, this effect can be observed by analyzing the weak term $g_{\theta}(\Tilde{\mathbf{x}}_{T-1}, c)$. It is less affected by accumulated denoising effects and better retains the motion characteristics expected from the weak term.

\begin{figure*}[!t]
\centering
\includegraphics[width=0.98\linewidth]{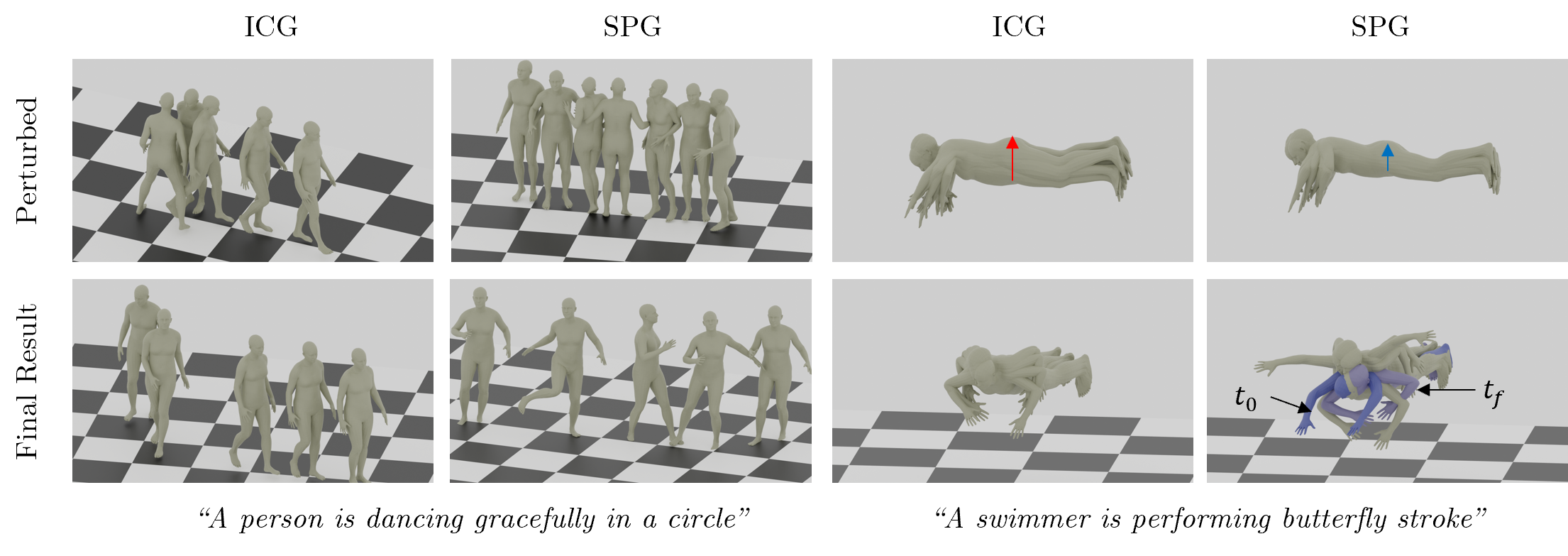}
   \caption{Comparison of the weak term $g_{\theta}(\mathbf{x}_{t}, c)$ at the early denosing step. (Left) ICG perturbation caused the loss of the semantic meaning of \textit{dancing}, resulting in a simple walking motion at the final denoising step. In contrast,  
   the weak term of SPG degraded the motion slightly while keeping the contents. SPG obtained more plausible final motion.  (Right) ICG perturbation leads to {unstable} sliding of the body, while SPG keeps {in-manifold} of the motion state as it applies the temporal smoothing. See the colored arrows. } 
\label{fig: weak_comparison}
\end{figure*}

The comparison results with ICG on the HumanML3D test set, evaluated over five repetitions, are summarized in Table \ref{tab: Alignment comparison}. As expected, the weak term of CFG, $g_{\theta}(\mathbf{x}_{T-1}, \phi)$, achieves the best FID since it is trained on arbitrary data within the motion dataset. SPG demonstrates a comparable FID to CFG, whereas ICG deviates more significantly from the original dataset. Furthermore, SPG best preserves the original content and motion intention, as reflected in the other two metrics. Qualitative analysis in Figure \ref{fig: weak_comparison} also indicates that SPG produces a more aligned weak term than ICG.
To approximately assess how well the intermediate output $\mathbf{x}_t$ remains on the state manifold under diffusion-guidance, the authors of \cite{huang2024constraineddiffusiontrustsampling} measured the \textit{trust region} using $\|\epsilon(\mathbf{x}_t, c)\|$. Similarly, we computed the average $\|\epsilon(\mathbf{x}_t, c)\|$ at each denoising step for the HumanML3D test set using the MDM model trained with 50 inference steps. As shown in Figure \ref{fig:manifold}, SPG consistently achieved a slightly lower $\|\epsilon(\mathbf{x}_t, c)\|$ value compared to other guidance methods.

\begin{table*}[h!]
\center
\begin{tabular}{lccccc}
\toprule
Guidance & FID ↓ & Matching Score ↓ & R precision (top  3) ↑ & Diversity → & MultiModality ↑ \\ \midrule
\multicolumn{6}{c}{\textbf{HumanML3D} } \\ \hline \rowcolor{gray!10}
           
Ground Truth    & 0.007 & 2.920  & 0.804 & 9.348    & -        \\  
\ding{55}       & 0.714 & 3.738  & 0.676 & {9.691} & 3.129        \\
ICG  & 0.607 & {3.534} & {0.716} & {9.775} & 2.766        \\
SG      & 0.457 & 3.745  & 0.672 & \textbf{9.504} & \textbf{3.300} \\ \rowcolor{green!10}
SPG           & \textbf{0.192} & {3.618} & {0.688} & \underline{9.542} & \underline{3.235} \\ \hline
CFG  & 0.532 & \underline{3.309} & \underline{0.752} & {10.001} & 2.172\\
CFG + ICG  & 0.527 & 3.318 &{0.750} & {9.962} & 2.139 \\
CFG + SG   & 0.426 & 3.347 & 0.740 & 9.889 & {2.353} \\ \rowcolor{green!10}
CFG + SPG& \underline{0.271} & \textbf{3.244} & \textbf{0.763} & 9.913 & {2.271}        \\ 
\midrule

\multicolumn{6}{c}{\textbf{KIT}} \\ \hline \rowcolor{gray!10}
Ground Truth    & 0.025   & 2.797         & 0.423     & 11.065    & -        \\   
\ding{55}       & 0.692 & {3.422} & 0.370 & 10.660 & 2.643        \\
ICG    & 0.595 & 3.227 & 0.708 & 10.707 & 2.400        \\
SG  &  0.688  & 3.518 & 0.680 & 10.701 & \textbf{2.845}        \\ \rowcolor{green!10}
SPG  & {0.528} & {3.421}          & {0.691} & {10.711} & \underline{2.718}        \\ \hline 

CFG             & 0.515 & \underline{3.110} & {0.723} & \textbf{10.852} & 1.898        \\
CFG + ICG   &  0.495        & \textbf{3.090}             & \textbf{0.733}                & \underline{10.769}             & 1.887        \\
CFG + SG  & 0.510        & 3.113             & \underline{0.732}                & 10.756             & 1.936        \\ \rowcolor{green!10}
CFG + SPG        & \textbf{0.411} & 3.168 & 0.720 & {10.758} & {1.928}        \\ 
\bottomrule
\end{tabular}
\\
\caption{Guidance comparison for T2M task in MDM}
\label{tab: mdm-t2m}

\end{table*}

\begin{table*}[]
\center
\begin{tabular}{lccccc}
\toprule
Guidance & FID ↓ & Matching Score ↓ & R precision (top 3) ↑& Diversity  → & MultiModality ↑ \\
\midrule
\multicolumn{6}{c}{\textbf{MotionCLR} } \\ \hline 

\ding{55} & {0.937} & {3.789} & 0.680 & {8.178} & 2.941 \\
ICG  & 0.975 & 3.822 & {0.686} & {8.242} & 2.853 \\
SG  & 0.990 & {3.796} & {0.681} & 8.1667 & \underline{2.955} \\

\rowcolor{green!10}  SPG & {0.563} & 3.823 & 0.679 & 8.008 & \textbf{2.989} \\
\midrule
CFG & \underline{0.278} & \underline{3.448} & \underline{0.756} & \underline{8.438} & 2.063 \\
CFG + ICG  & {0.280} & {3.455} & \textbf{0.757} & \underline{8.382} & 2.029 \\
CFG + SG  & 0.291 & \textbf{3.444} & \underline{0.756} & \textbf{8.438} & {2.064} \\
\rowcolor{green!10}  CFG + SPG  & \textbf{0.231} & 3.486 & 0.744 & 8.279 & {2.108} \\
\midrule

\multicolumn{6}{c}{\textbf{GMD} } \\ \hline \rowcolor{gray!10}

\rowcolor{gray!10} Ground Truth & 0.001 & 2.967 & 0.797 & 9.594 & - \\
\ding{55} & 0.358 & 5.782 & 0.568 & \textbf{9.596} & 3.571 \\
ICG  & 0.339 & 5.803 & 0.561 & \underline{9.521} & 3.591 \\
SG  & 0.391 & 5.974 & 0.547 & 9.301 & \textbf{3.781} \\
\rowcolor{green!10} SPG  & \underline{0.197} & 5.645 & 0.583 & 9.466 & \underline{3.655} \\
\midrule
CFG  & 0.287 & 5.276 & \underline{0.655} & 10.037 & 2.472 \\
CFG + ICG  & 0.229 & \underline{5.259} & 0.650 & 9.932 & 2.439 \\
CFG + SG & 0.399 & 5.469 & 0.626 & 9.711 & 2.686 \\
\rowcolor{green!10} CFG + SPG  & \textbf{0.176} & \textbf{5.165} & \textbf{0.667} & 9.791 & 2.532 \\
\bottomrule

\end{tabular}
\caption{Guidance comparison for GMD and MotionCLR for text-to-motion HumanML3D testset.}
\label{tab: motionclr-gmd-t2m}
\end{table*}

\begin{table*}[]
\center
\begin{tabular}{lccccc}
\toprule
            & {FID}   ↓     & {Diversity}  →      & {Control L2} ↓ & {Skating Ratio}  ↓ & {R Precision (top 3) ↑ } \\ \midrule
\multicolumn{6}{c}{\textbf{Spatial constraint}} \\ \midrule
\rowcolor{gray!10} Ground Truth                        & 0.007             & 9.494            & -    & -    & 0.506            \\

\ding{55}                    & 0.651              & 9.169                    & \underline{0.052}    & \underline{0.074}       & \underline{0.371}             \\
ICG             & 0.968              & 8.972                    & 0.068                & 0.082                   & 0.266                        \\
 SG        & \underline{0.624}  & \textbf{9.322}           & 0.053                & 0.079                   & 0.344                        \\
\rowcolor{green!10} SPG  & \textbf{0.435}     & \underline{9.193}        & \textbf{0.050}       & \textbf{0.067}          & \textbf{0.398}               \\ \midrule

\multicolumn{6}{c}{\textbf{Spatial constraint + prompt}} \\ \midrule

\ding{55} & {0.216}  & \textbf{9.576}        & \textbf{0.042}       & {0.072}       & {0.627}             \\
ICG   & 0.619 & 8.847                    & 0.063                & 0.073                   & 0.616                        \\
SG   & {0.164}  & 9.376  & 0.047                & 0.086                   & 0.630                        \\
\rowcolor{green!10} SPG            & \textbf{0.112}     & \underline{9.639}           & \textbf{0.042}       & \underline{0.060}          & {0.637}               \\
\hline
CFG                       & {0.158}  & {9.766}        & \underline{0.044}       & {0.063}          & {0.665}             \\
CFG +  ICG          & 0.285              & 9.244                    & 0.054                & 0.065                   & {0.683}               \\
CFG  + SG      & {0.145}     & {9.691}        & {0.045}    & {0.065}       & \textbf{0.674}                        \\
\rowcolor{green!10} CFG + SPG  & \underline{0.130}              & {9.819}           & 0.047                & \textbf{0.053}                   & \underline{0.672}                        \\ \bottomrule

\end{tabular}
\\
\caption{Guidance comparison for OmniControl (pelvis trajectory following) \cite{omnicontrol}}
\label{tab:omnicontrol}
\end{table*}

\begin{table*}[h!]
\center
\begin{tabular}{lcccc}
\toprule 
\multicolumn{5}{c}{\textbf{A2M task}} \\
\midrule
Guidance & Accuracy ↑ & Diversity → & FID  ↓ & Multimodality ↑ \\
\midrule
\multicolumn{5}{c}{HumanAact12} \\ 
\midrule
\rowcolor{gray!10} Ground Truth & 0.994 & 6.879 & 0.001 & 2.583 \\
\ding{55} & \underline{0.985} & \textbf{6.880} & 0.100 & 2.529 \\
ICG& \textbf{0.986} & \underline{6.875} & 0.102 & 2.526 \\
SG & 0.983 & 6.872 & \underline{0.099} & \textbf{2.556} \\
\rowcolor{green!10} SPG & \underline{0.985} & 6.864 & \textbf{0.094} & \underline{2.552} \\
% \end{tabular}
% \caption{Guidance comparison for A2M task in MDM \cite{mdm}}
% \label{tab:mdm-a2m}
% \end{table*}
\midrule
\multicolumn{5}{c}{UESTC} \\ 
\midrule
\rowcolor{gray!10} Ground Truth & 0.988 & 33.34 & 2.79 & 14.16 \\
\ding{55} & 0.947 & 32.96 & 12.94 & 14.26 \\
ICG& 0.947 & \underline{32.97} & 13.01 & \textbf{14.40} \\
SG & 0.944 & 32.96 & \underline{12.77} & 14.35 \\
\rowcolor{green!10} SPG & 0.947 & \textbf{33.00} & \textbf{12.12} & \textbf{14.40} \\
% \begin{table*}[h!]
% \centering
% \begin{tabular}{lcccc}
\toprule
\multicolumn{5}{c}{\textbf{Unconstrained generation}} \\
\midrule
Guidance & FID ↓ & KID ↓ & Precision ↑ & Recall ↑ \\
\midrule
\ding{55} & 31.261 & 0.368 & \underline{0.702} & \textbf{0.691} \\
 SG  & \underline{28.015} & \textbf{0.276} & 0.697 & 0.607 \\
\rowcolor{green!10} SPG & \textbf{26.783} & \underline{0.281} & \textbf{0.704} & \underline{0.666} \\
\bottomrule
\end{tabular}
\caption{Guidance comparison for MDM models not trained on CFG}
\label{tab:mdm-uncond}
\end{table*}

\section{Results}
This section presents comprehensive results across various tasks and model architectures, comparing SPG with other state-of-the-art methods on model-agnostic sampling guidance: ICG \cite{icg} and SG \cite{sg}. 
The supplementary material includes detailed setups for experiments.
% All experiments were conducted on Nvidia H100 GPUs and repeated five times for each setting. In the case of unconditional synthesis and A2M, we repeated the process 20 times. 
% For all the sampling methods, parameters were selected based on the lowest FID score and can be found in the supplementary material. In the tables below, the best result is highlighted in {bold}, while the second-best is {underlined}.

\subsection{Transformer architecture}
As one of the representative networks, MDM \cite{mdm} utilizes a transformer-only structure by integrating trajectory embeddings $\mathbf{x}_t$ and prompts $c$ into a token sequence. Since the authors provided T2M models trained on both HumanML3D and KIT, we conducted experiments on both datasets.
As summarized in Table \ref{tab: mdm-t2m}, SPG outperformed CFG in terms of FID and diversity, demonstrating the advantages of weak models, as observed in prior studies \cite{pag, autoguidance} in the image generation. Moreover, SPG can be combined with CFG to further enhance its performance including R-precision, showcasing its complementarity with CFG, a notable strength of weak models \cite{sag}. Whether  CFG was combined or not, SPG showed the best results in the fidelity among all guidance methods. Moreover, SPG consistently outperformed other guidances or the baseline in terms of FID, including the A2M task and unconstrained generation task as shown in Table \ref{tab:mdm-uncond}.

\subsection{UNet architecture}
MotionCLR \cite{motionclr} employs a UNet architecture with self- and cross-attention blocks, closely resembling image generation networks \cite{sd, podell2023sdxl}. As shown in the first four rows of Table \ref{tab: motionclr-gmd-t2m}, only SPG improved FID across both configurations (with and without CFG), whereas other model-agnostic guidance methods failed to enhance the denoising process.
In contrast, GMD \cite{gmd} does not incorporate attention mechanism and instead employs a pure UNet. The results on guidance comparison are summarized in Table \ref{tab: motionclr-gmd-t2m}. In the T2M task, SPG improved both fidelity and multi-modality, and its trend is consistent with MDM and MotionCLR. Overall, combining CFG and SPG yielded the best performance.

\subsection{Adaptor structure}
To investigate whether SPG can improve the diffusion networks having additional guidance network, we test SPG to Omnicontrol \cite{omnicontrol}, where realism guidance injects intermediate outputs into transformer blocks. The results are presented in Table \ref{tab:omnicontrol}.
We conducted a pelvis-control experiment, where the prompt could either be provided or omitted. In the latter case, only a spatial constraint was applied without CFG. Here, SPG achieved the best performance. When both spatial constraint and prompt were given, SPG attained the best FID, while the combination of CFG and SPG resulted in the highest R-precision. Notably, SPG has consistently  reduced foot-skating. 

\subsection{Models not trained for CFG}
Among the introduced networks, MDM \cite{mdm} explored A2M and unconstrained generation, both of which do not have an additional training on the unconditional term, rendering CFG inapplicable. In these cases, the weak model guidance is the only option when test-time modification is considered.
As summarized in Table \ref{tab:mdm-uncond}, in the A2M task, SPG achieves the best FID and the second-best results for accuracy and multi-modality for HumanAct12 \cite{guo2020action2motion}, and topped all the metrics in UESTC \cite{ji2019largescalevaryingviewrgbdaction} dataset. 
For unconditional generation, we excluded ICG as it perturbs conditional embeddings, which is not applicable in this setting. As the result shows, SPG outperformed SG. 

\section{Conclusion }
This paper introduced Smooth Perturbation Guidance (SPG) as an aligned weak-model approach specifically designed for motion generation. As a model-agnostic method, SPG achieved state-of-the-art fidelity across a diverse range of model architectures. By leveraging the advantages of weak models originally developed for image generation, SPG enhanced output quality without sacrificing diversity and demonstrated complementarity with Classifier-Free Guidance (CFG).

SPG has certain limitations. It requires an additional model evaluation, which cannot be processed in a single batch like CFG. We attached computation tables in the supplementary. Moreover, while temporal smoothing effectively reduces noise, it can sometimes introduce abrupt changes in motion outputs. This occurs when the weak term unintentionally acts as negative guidance against slow movements. Refer to the supplementary materials for related analysis. 
Future work will focus on refining SPG to address these limitations while preserving its broad applicability across different motion generation tasks.

{
    \small
    \bibliographystyle{ieeenat_fullname}
    \bibliography{main}
}

\section{Supplementary Material}

\subsection{Experiment setup}

All experiments were conducted on Nvidia H100 GPUs and repeated five times for each setting. In the case of unconditional synthesis and A2M, we repeated the process 20 times. 
For all the sampling methods, parameters were selected based on the lowest FID score. 
It is important to note that the interpretation of the scale parameter varies across methods, as it follows the formulation:

\[
\hat{\mathbf{x}}_{0}= g_{\theta}(\mathbf{x}_t, c)  + s \big(g_{\theta}(\mathbf{x}_t, c) - \Tilde{g}_{\theta}(\mathbf{x}_t, c) \big).
\]
For instance, the CFG scale \( s \) in the MDM paper is one unit larger than in our setup. In the case of SG, \( \delta t \) corresponds to the \textit{Dynamic Shift Scale} as defined in its original paper.  
For rows that include CFG, we set the CFG scale to \( 1.5 \).

\begin{table}[t!]
    \centering
    \renewcommand{\arraystretch}{1.2}
    \begin{tabular}{l c c c}
        \toprule
        \textbf{Guidance} & None & CFG & SPG \\
        \midrule
        \textbf{Computation [s]} & 0.2758 & 0.4124 & 0.5581 \\
        \bottomrule
    \end{tabular}
    \caption{Average calculation times (repeated 10 times)}
    \label{tab:guidance-averages}
\end{table}

\begin{table*}[h!]
    \centering
    \renewcommand{\arraystretch}{1.2}
    \setlength{\tabcolsep}{4pt} % Adjust column spacing
    \small % Reduce font size for compactness
    
    \begin{adjustbox}{max width=\textwidth}
    \begin{tabular}{l l l l}
        \toprule
        \textbf{Task} & \textbf{Dataset} & \textbf{Guidance} & \textbf{Parameters} \\
        \midrule
        
        \multirow{7}{*}{\textbf{T2M}} 
        & \multirow{7}{*}{\textbf{MDM (HumanML3D)}} 
        & ICG & \( (s=1.5) \)  \\
        & & SG & \( (s=0.7, \delta t=20) \)  \\
        & & SPG & \( (s=0.3, k=5) \)    \\
        & & CFG & \( (s=1.5) \)  \\
        & & CFG + ICG & \( (s=1.0) \) \\
        & & CFG + SG & \( (s=1.0, \delta t=20) \) \\
        & & CFG + SPG & \( (s=0.3, k=5) \) \\
        \midrule

        & \multirow{7}{*}{\textbf{MDM (KIT)}} 
        & ICG & \( (s=1.2) \)   \\
        & & SG & \( (s=0.5, \delta t=20) \)  \\
        & & SPG & \( (s=0.1, k=7) \)  \\
        & & CFG  & \( (s=1.5) \)  \\
        & & CFG + ICG & \( (s=0.5) \)  \\
        & & CFG + SG & \( (s=0.1, \delta t=40) \)  \\
        & & CFG + SPG  & \( (s=0.1, k=7) \) \\
        \midrule
        
        & \multirow{6}{*}{\textbf{MotionCLR}} 
        & ICG & \( (s=0.3) \) \\
        & & SG & \( (s=0.1, \delta t=40) \)  \\
        & & SPG & \( (s=0.1, k=11) \) \\
        & & CFG & \( (s=1.5) \) \\
        & & CFG + ICG & \( (s=0.3) \) \\
        & & CFG + SG & \( (s=0.1, \delta t=40) \) \\
        & & CFG + SPG & \( (s=0.1, k=9) \) \\
        \midrule
        
        & \multirow{7}{*}{\textbf{GMD}} 
        & ICG & \( (s=0.5) \) \\
        & & SG & \( (s=0.5, \delta t=40) \) \\
        & & SPG & \( (s=0.1, k=7) \) \\
        & & CFG & \( (s=1.5) \) \\
        & & CFG + ICG & \( (s=0.5) \) \\
        & & CFG + SG & \( (s=0.75, \delta t=40) \) \\
        & & CFG + SPG & \( (s=0.25, k=3) \) \\
        \midrule
        
        \multirow{6}{*}{\textbf{Spatial Constraint}} 
        & \multirow{6}{*}{\textbf{OmniControl}} 
        & ICG & \( (s=0.5) \) \\
        & & SG & \( (s=0.3, \delta t=20) \)  \\
        & & SPG & \( (s=0.1, k=3) \) \\
        & & ICG (Spatial + Prompt) & \( (s=0.7) \) \\
        & & SG (Spatial + Prompt) & \( (s=1.0, \delta t=20) \)  \\
        & & SPG (Spatial + Prompt) & \( (s=0.2, k=3) \)  \\
        \midrule
        
        \multirow{4}{*}{\textbf{A2M}} 
        & \multirow{4}{*}{\textbf{MDM}} 
        & ICG & \( (s=0.5) \) \\
        & & SG & \( (s=0.3, \delta t=40) \) \\
        & & SPG & \( (s=0.2, k=3) \) \\
        \midrule
        
        \multirow{2}{*}{\textbf{Unconstrained Gen.}} 
        & \multirow{2}{*}{\textbf{MDM}} 
        & SG & \( (s=0.7, \delta t=20) \) \\
        & & SPG & \( (s=0.7, k=5) \) \\
        \bottomrule
    \end{tabular}
    \end{adjustbox}

    \caption{Comparison of guidance methods across various datasets and tasks.}
    \label{tab:all-guidance}
\end{table*}

\subsection{Computation}
For a single batch, the computation time was measured in Nvidia H100 GPU and we used MDM official implementation. See Table. \ref{tab:guidance-averages}

\subsection{Increased derivatives when SPG is applied}
SPG built the negative term $g_{\theta}(\Tilde{\mathbf{x}}_{t}, c)$ by temporal smoothing, degrading motion details. At the same time, this negative term can also correspond to a slowdown of the original motion.
When applied as negative guidance during sampling, SPG can lead to increased velocity or acceleration in the generated motion.
To evaluate this side effect, we conducted experiments using 50 prompts that involve large movements, testing SPG on MDM trained on HumanML3D. Although SPG can generate more accurate motion with higher fidelity, it may also introduce abrupt transitions with increased acceleration.

\begin{figure}[h]
\centering
\includegraphics[width=0.9\linewidth]{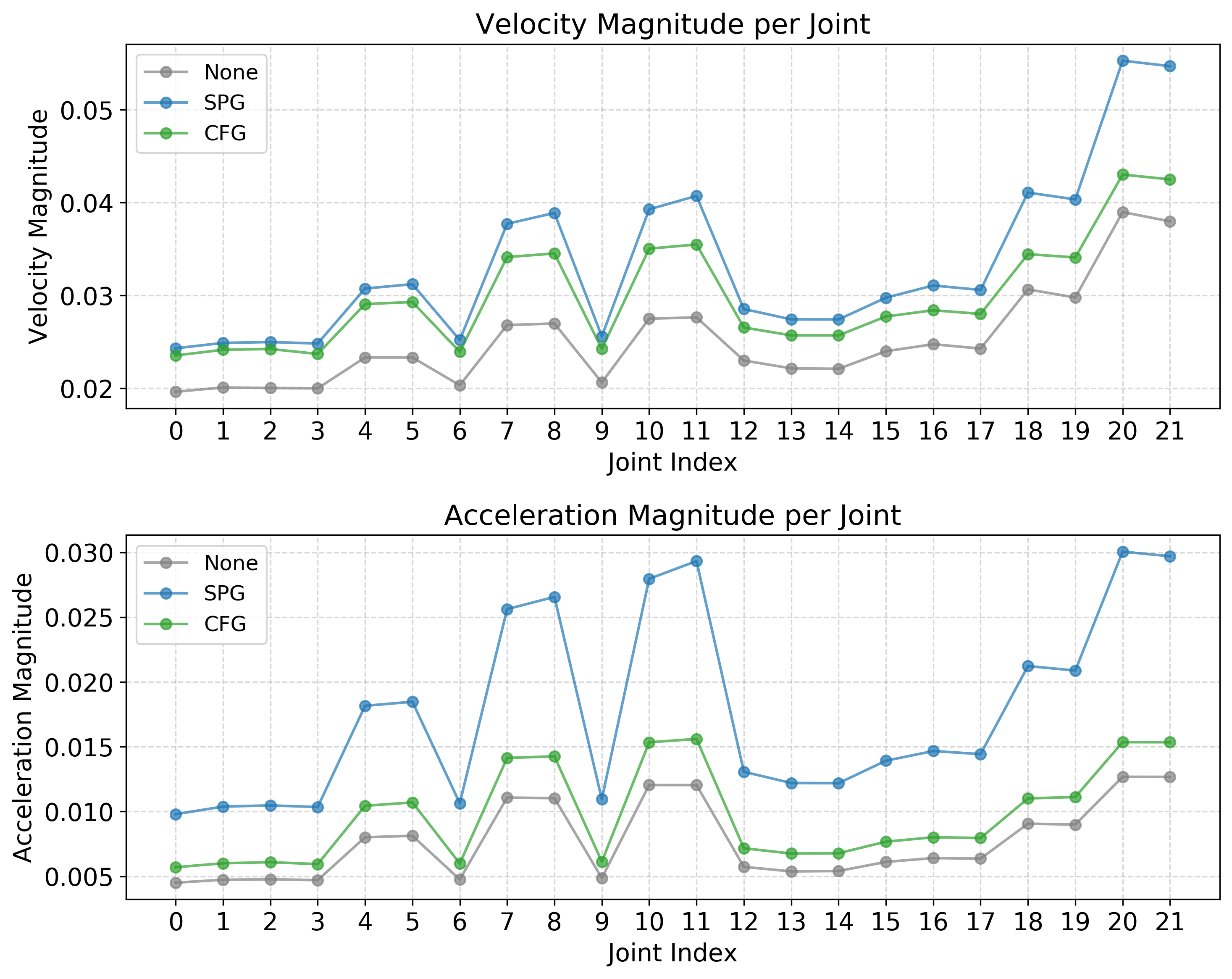}
   \caption{Speed (top) and acceleration (bottom) comparsion for guidance methods. The joint indexing is based on SMPL 22 joints of HumanML3D.}

\label{fig:long_1}
\end{figure}

\subsection{Comparison of smoothing axis}
\begin{figure}[h]
\centering
\includegraphics[width=0.99\linewidth]{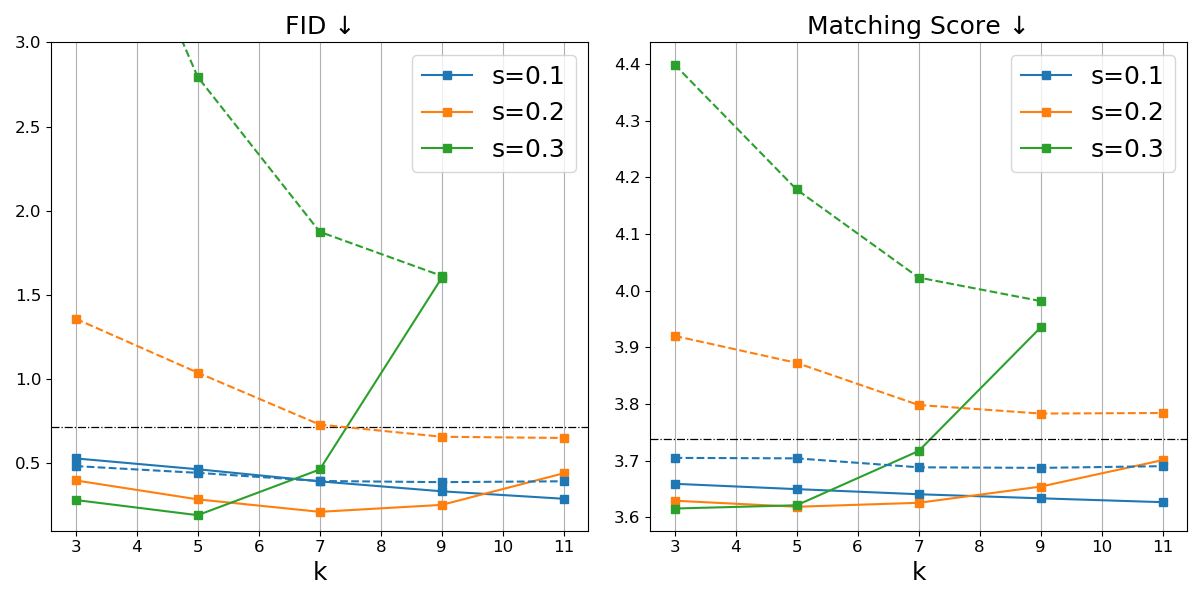}
\includegraphics[width=0.99\linewidth]{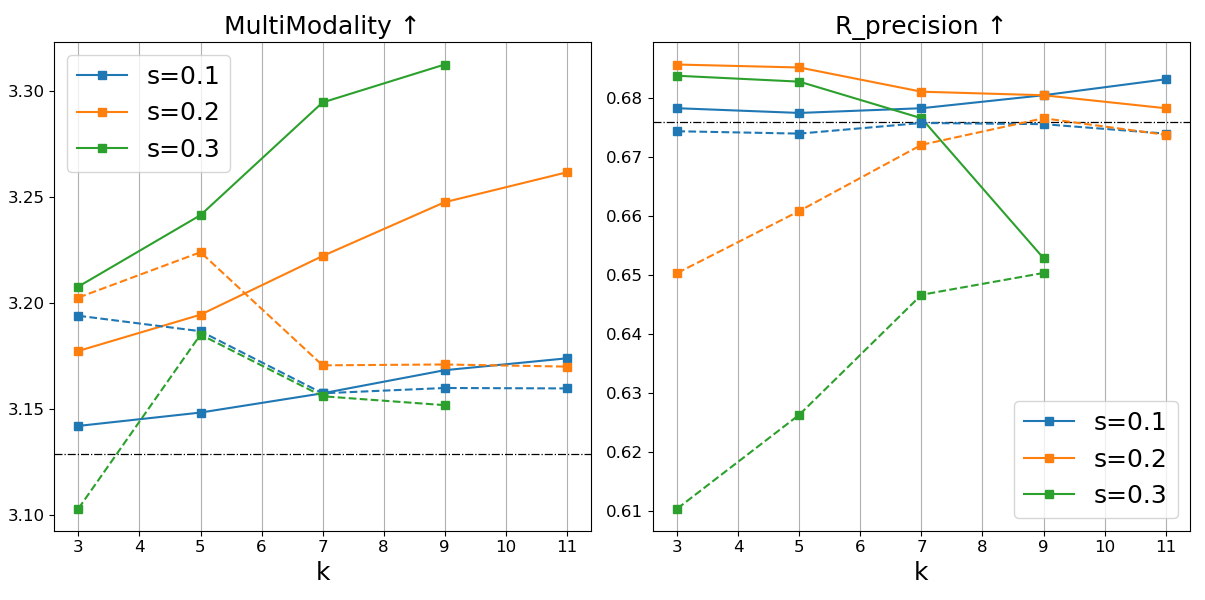}
\caption{Performance for various SPG scales $s$ and kernel sizes $k$. Evaluations were conducted on the HumanML3D test set using the T2M model of MDM without CFG. Black dash denotes the baseline without any weak model guidances.
Colored dashed lines were obtained from the original SAG implementation, where both axes are smoothed. Solid lines correspond to SPG.} 
\label{fig: k scale sspg}
\end{figure}

The original SAG implementation applied blurring to the input $\Tilde{\mathbf{x}}_{t}$ in both axis. As expected, smoothing along the spatial axis leads to an out-of-manifold distribution relative to the motion dataset. Consequently, guidance applied to both axes performed even worse than the baseline (black dashed line).


\end{document}